\documentclass{article}

\usepackage[nonatbib, preprint]{neurips_2025}

\usepackage[utf8]{inputenc} 
\usepackage[T1]{fontenc}    
\usepackage{hyperref}       
\usepackage{url}            
\usepackage{booktabs}       
\usepackage{amsfonts}       
\usepackage{amssymb}        
\usepackage{nicefrac}       
\usepackage{microtype}      
\usepackage{amsmath}
\usepackage{graphicx}

\usepackage{nicematrix}
\usepackage{booktabs}
\usepackage{algorithm}
\usepackage{algpseudocode}
\usepackage{multirow}
\usepackage{hhline}
\usepackage{cellspace}             
\usepackage{pifont}
\usepackage{graphicx}
\usepackage{amsmath}
\usepackage{multirow}
\usepackage{makecell}
\usepackage{pifont}
\usepackage{placeins}
\usepackage{hyperref}
\usepackage{url}
\usepackage{marvosym}
\usepackage{dsfont}
\usepackage{fdsymbol}
\usepackage{tikz}
\usepackage{colortbl} 
\usepackage{nicefrac}
\usepackage{marvosym}
\newcommand{\best}[1]{\textbf{\textcolor{red}{#1}}}
\newcommand{\secondbest}[1]{\underline{\textcolor{blue}{#1}}}
\newcommand\shline{\specialrule{0.8pt}{0pt}{0pt}}

\definecolor{OursColor}{HTML}{FFFFCC}  
\definecolor{myIDBcolor}{HTML}{FFF5F0}
\definecolor{myCCDBcolor}{HTML}{F5FFF0}  

\title{MIND-Edit: MLLM Insight-Driven Editing via Language-Vision Projection}

\author{\hspace{-9pt}
Shuyu Wang\textsuperscript{\textcolor{black}{$\clubsuit$}1},
Weiqi Li\textsuperscript{\textcolor{black}{$\clubsuit$}1},
Qian Wang\textsuperscript{1},
Shijie Zhao\textsuperscript{2}, 
Jian Zhang\textsuperscript{1~\Letter}\\
\textsuperscript{1} School of Electronic and Computer Engineering, Peking University\\
\textsuperscript{2} ByteDance Inc.\\
}

\begin{document}

\maketitle
\renewcommand*{\thefootnote}{$\clubsuit$}
\footnotetext[1]{Equal contribution. \Letter: Corresponding author, zhangjian.sz@pku.edu.cn.}
\renewcommand*{\thefootnote}{\arabic{footnote}}

\begin{abstract}
Recent advances in AI-generated content (AIGC) have significantly accelerated image editing techniques, driving increasing demand for diverse and fine-grained edits. Despite these advances, existing image editing methods still face challenges in achieving high precision and semantic accuracy in complex scenarios. Recent studies address this issue by incorporating multimodal large language models (MLLMs) into image editing pipelines. However, current MLLM-based methods mainly rely on interpreting textual instructions, leaving the intrinsic visual understanding of large models largely unexplored, thus resulting in insufficient alignment between textual semantics and visual outcomes. To overcome these limitations, we propose \textbf{MIND‑Edit}, an end‑to‑end image‑editing framework integrating pretrained diffusion model with MLLM. MIND‑Edit introduces two complementary strategies: (1) a text instruction optimization strategy that clarifies ambiguous user instructions based on semantic reasoning from the MLLM,  and (2) an MLLM insight-driven editing strategy that explicitly leverages the intrinsic visual understanding capability of the MLLM to infer editing intent and guide the diffusion process via generated visual embeddings. Furthermore, we propose a joint training approach to effectively integrate both strategies, allowing them to reinforce each other for more accurate instruction interpretation and visually coherent edits aligned with user intent. Extensive experiments demonstrate that MIND-Edit outperforms state-of-the-art image editing methods in both quantitative metrics and visual quality, particularly under complex and challenging scenarios. 
\end{abstract}

\section{Introduction}\label{sec:Intro}
Image editing is a process of modifying, enhancing, compositing, or creating images to achieve a desired visual effect, which plays a central role in many computer vision applications. With the rapid advancement of artificial intelligence, deep learning techniques such as generative adversarial networks (GANs)~\cite{goodfellow2014generative,karras2019style,karras2020analyzing} and variational autoencoders (VAEs)~\cite{kingma2013auto} have demonstrated impressive capabilities in image generation and editing tasks. Recently, diffusion models~\cite{ho2020denoising,stablediffusion,podell2023sdxldiffusionmodels,saharia2022photorealisticdiffusionmodels} enable the generation of high-quality and diverse images that capture complex structures and fine details, providing a powerful foundation for image editing. 
Existing image editing methods can be categorized into three types based on their interaction paradigms and control mechanisms: (1) text-driven methods~\cite{tumanyan2023plugt2iedit,brooks2023instructpix2pix,sheynin2024emutextedit1,dao2024swiftbrusht2idiffusionmodels,liu2025step1x-edit2iedit}, which rely on pre-trained diffusion models to interpret natural language instructions; (2) drag-based methods~\cite{pan2023drag,shi2024dragdiffusion,mou2024dragondiffusion,liu2024drag,zhang2024gooddrag}, which allow users to manipulate specific image features with pixel-level control; and (3) visually guided techniques, which use segmentation masks~\cite{zhao2024ultraeditsegedit,avrahami2022blendedsegedit}, layout bounding boxes~\cite{chen2024trainingbox1,wang2024instancediffusionbox2,song2025insertrefedit,chen2024MimicBrushrefedit}, or reference images~\cite{li2024brusheditimageedit,chen2024anydoorimagecontent} to guide the editing process. However, text-driven methods suffer from inherent ambiguity and often fail to align with user intent; drag-based methods are time-consuming and lack scalability for complex edits, and visually guided techniques heavily depend on precise user-supplied conditions, which are frequently imprecise. More importantly, all these methods share a problematic assumption: that user input accurately and completely conveys editing intent. In practice, users often express their intentions imprecisely, yet current systems treat these imperfect inputs as entirely accurate, causing errors to propagate and compound throughout the generation pipeline. Therefore, accurately inferring the user's true editing intent is critical for achieving high-quality editing results.

Multi-modal large language models (MLLMs)~\cite{liu2023visual,llavaonevision}, which possess extensive world knowledge and exhibit strong capabilities in semantic understanding and visual reasoning, have recently been integrated into image editing frameworks to effectively understand user instructions. Existing MLLM-based approaches typically fall into two categories. The first category treats the MLLM as a "commander"~\cite{ju2024imagemllmcommander,wang2024genartist,gu2024multicommander,wu2024visionllmcommander}, interpreting user instructions and delegating tasks to external editing tools or decomposing them into simpler operations. However, this reliance on external modules often results in fragmented execution pipelines, compromising end-to-end consistency and editing precision. The second category employs MLLMs primarily as text encoders~\cite{huang2024smartedit,wang2024flexedit,koh2023generatingtextmllmedit,fu2023guidingmllmeditmgietext,ju2024brushnet}, translating complex instructions into prompts for diffusion models. These approaches largely ignore the MLLMs' inherent visual reasoning capability, causing a semantic gap between textual prompts and actual user intent, especially in ambiguous or detail-sensitive editing scenarios. Similarly, the recent stroke-based method MagicQuill~\cite{liu2024magicquill} converts freeform user-drawn strokes into simple object-level nouns, which may lack sufficient expressive power for precise edits. Despite this meaningful progress, existing methods still struggle to effectively translate users' imprecise inputs into accurate visual outcomes, especially in scenarios requiring deep visual-semantic alignment.

Recent advances in unified visual understanding-generation frameworks~\cite{xie2024showunify,wang2024emu3unify,team2024chameleonunify,tong2024metamorph,pan2025transfer} have leveraged the visual reasoning capability intrinsic to MLLMs, allowing these models to implicitly interpret complex generative instructions within integrated understanding-generation paradigms. For example, MetaMorph~\cite{tong2024metamorph} and MetaQueries~\cite{pan2025transfer} specifically decouple the understanding and generation stages, using MLLMs first to produce visual representations and subsequently injecting these into diffusion models to improve image generation quality.  These works indicate that MLLMs, while originally designed for understanding, exhibit notable generative capabilities under limited supervision and suggest that jointly leveraging understanding and generation can mutually enhance performance during training. Motivated by these insights, we propose extending this paradigm to image editing, arguing it provides three distinctive advantages: \textbf{(1)} the MLLMs' visual understanding capability naturally disambiguates vague or imprecise textual instructions provided by users;
\textbf{(2)} the visual representations produced by MLLMs effectively guide downstream diffusion models toward edits closely aligned to users' intent; \textbf{(3)} jointly optimizing the textual and visual outputs enhances coherence and synergy between modalities, further improving the precision of editing outcomes.

Building on this insight, we introduce \textbf{MIND-Edit}, an end-to-end image editing framework that integrates a diffusion model with an MLLM. MIND-Edit enhances editing accuracy and controllability through two complementary strategies: \textbf{(1)} an instruction optimization strategy leveraging the MLLM’s semantic reasoning and visual comprehension to transform ambiguous user instructions into clear, executable directives; and \textbf{(2)} an MLLM insight-driven editing strategy explicitly utilizing the MLLM’s visual understanding capability, generating visual representations to accurately infer the intended edits and directly guide the diffusion process. A joint training approach is adopted to further encourage mutual reinforcement of semantic and visual capabilities, enabling the MLLM to simultaneously produce refined textual instructions and corresponding visual representations in a unified manner. By integrating these strategies and the joint training approach, MIND-Edit effectively bridges the gap between imprecise user language input and high-quality visual outcomes. Extensive experiments demonstrate that MIND-Edit consistently outperforms state-of-the-art image editing methods in terms of quantitative metrics and visual quality, especially when handling complex and challenging editing scenarios. In summary, our contributions are fourfold:

\ding{113} We propose MIND-Edit, a plug-and-play, end-to-end framework for accurate and controllable image editing, integrating a diffusion model with a multimodal large language model (MLLM) to significantly improve editing accuracy and quality.

\ding{113} We develop an MLLM insight-driven image editing strategy, explicitly leveraging the MLLM's visual understanding capability to predict visual representations of desired edits, which directly guide the diffusion model toward results closely aligned with user intent.

\ding{113} We introduce a joint training approach, enabling the MLLM to simultaneously generate refined textual instructions and corresponding visual representations. This strategy strengthens mutual enhancement between textual and visual modalities, further improving editing performance.

\ding{113} We conduct extensive quantitative and qualitative experiments across multiple datasets, demonstrating consistent superiority over state-of-the-art image editing techniques. Detailed ablation studies further validate the effectiveness of each component.

\section{Related Works}\label{sec:Realted}
\subsection{Image Editing}
With the rapid development of deep learning technologies, numerous image editing methods have been proposed and widely applied across various scenarios. Each paradigm addresses specific aspects of image editing while exhibiting distinct limitations.
Recent deep learning approaches enable precise pixel-level manipulation through point-dragging interactions~\cite{liu2024drag,pan2023drag,zhang2024gooddrag,li2024omnidrag}. DragGAN \cite{pan2023drag} pioneered a drag-based interaction paradigm for image editing, but its performance is limited by the generative capacity of GANs. DragDiffusion \cite{shi2024dragdiffusion} transferred this interaction paradigm to diffusion models. DragonDiffusion \cite{mou2024dragondiffusion} and DiffEditor \cite{mou2024diffeditor} further formalized the editing process as energy-guided latent optimization. 
Visually guided methods leverage spatial conditions such as masks, edges, or poses to guide editing~\cite{zhao2024ultraeditsegedit,avrahami2022blendedsegedit,narasimhaswamy2024handiffuserposetextt2i,li2024adversarialseg,wang2024360dvd,li2024resvr}. ControlNet~\cite{controlnet} introduced a general framework for incorporating spatial priors into diffusion models. This approach was further extended by T2I-Adapter~\cite{mou2024t2i}, which adopts a more lightweight and flexible design. While effective, these methods rely heavily on the precision and quality of user-provided visual inputs.
Text-driven methods allow users to specify edits through natural language prompts. InstructPix2Pix~\cite{brooks2023instructpix2pix} pioneered this direction with paired image-text supervision. Follow-up methods, including MagicBrush~\cite{zhang2023magicbrush}, AnyEdit~\cite{yu2024anyedit}, Prompt-to-Prompt~\cite{hertz2022prompt}, RAG-diffusion~\cite{chen2024regionattn}, and TextCrafter~\cite{du2025textcrafterattn}, improve alignment through attention manipulation or mask guidance. Recent methods~\cite{garibi2024renoisetextt2ifast,deutch2024turboeditt2itextfast,starodubcev2024invertibletextt2ifast} improve inference speed while maintaining high visual fidelity. However, these methods still face challenges due to the inherent semantic gap between textual instructions and visual interpretation.

\subsection{MLLMs for Image Editing}
Multimodal large language models (MLLMs) have recently shown great promise in adapting to image generation and editing tasks~\cite{ju2024imagemllmcommander,fu2023guidingmllmeditmgietext,koh2023generatingtextmllmedit,liu2024magicquill,huang2024smartedit,wang2024flexedit,li2025q}. Studies such as SmartEdit~\cite{huang2024smartedit} and FreeEdit~\cite{he2024freeeditllm} fine-tuned an MLLM, thereby adapting it to image editing tasks. FlexEdit~\cite{wang2024flexedit} integrates an MLLM to enable joint parsing of image content, masks, and textual instructions. GenArtist and VisionLLM~\cite{wang2024genartist,wu2024visionllmcommander} treat the MLLM as an intelligent agent capable of decomposing complex tasks, selecting and invoking appropriate tools. MagicQuill~\cite{liu2024magicquill} simplifies the editing process into an innovative "drawing-as-editing" paradigm through the use of an MLLM.
While these studies represent meaningful progress, they primarily focus on the MLLMs' ability to process and decompose textual instructions. In contrast, their strong inherent capability for visual reasoning remains underutilized.
Recent work has shown that the visual understanding capabilities of MLLMs can be effectively leveraged to guide diffusion-based image generation. For example, MetaQueries~\cite{pan2025transfer}, building on ideas from MetaMorph~\cite{tong2024metamorph}, introduces a connector that transforms visual tokens generated by an MLLM into a form compatible with diffusion models. This establishes a unified paradigm that integrates the interpretability of MLLMs with the generative power of diffusion models. Together, these studies demonstrate the potential of MLLM-driven visual representations to enhance image generation quality through direct integration with diffusion architectures.

\begin{figure}[!t]
    \centering
    \includegraphics[width=1\textwidth]{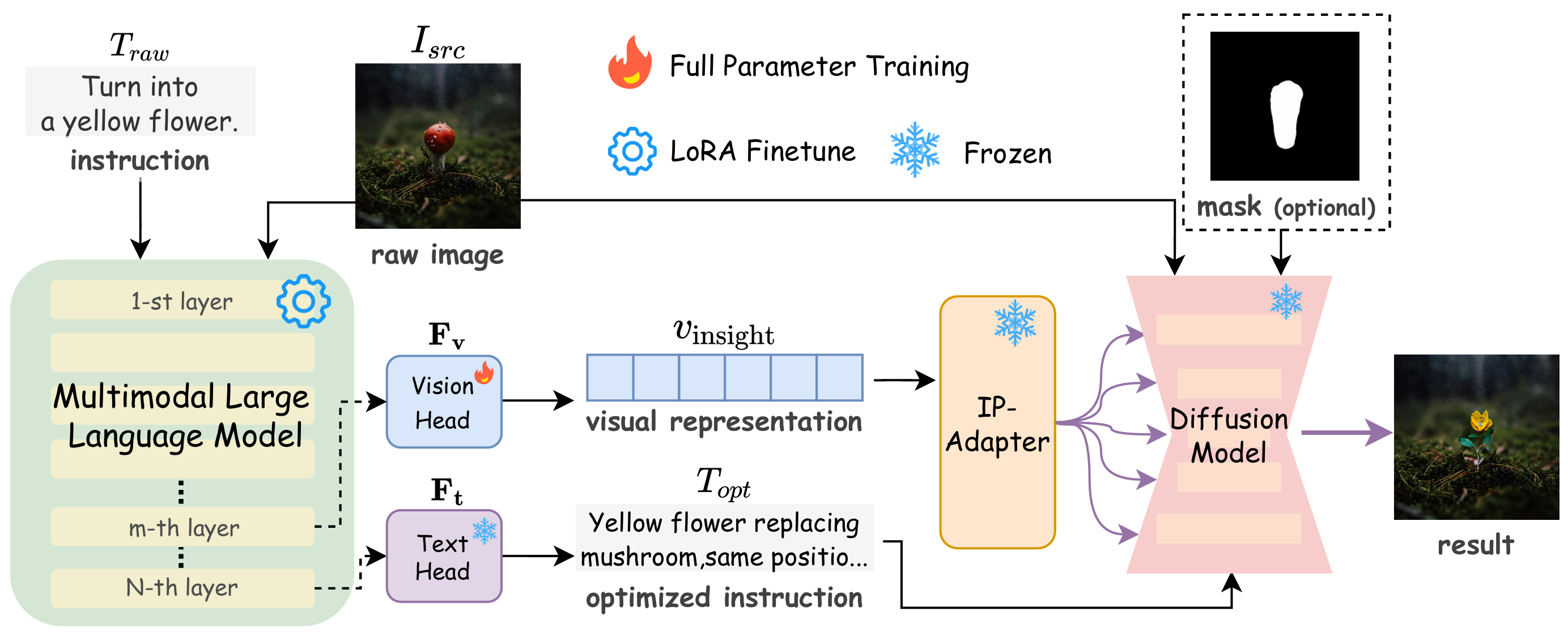}
    \caption{\textbf{Overview of the proposed MIND-Edit framework.} MIND-Edit takes text instructions, original images, and optional editing masks as inputs. It integrates an instruction optimization strategy and an MLLM insight-driven image editing strategy, jointly optimizing instructions and generating visual representations to guide the diffusion model in creating semantically accurate edited images.}
    \label{fig:overview}
\end{figure}

\section{Methodology}\label{sec:method}
\subsection{Preliminaries}

\textbf{Text-guided Image Editing.} 
The goal of text-guided image editing is to apply specific modifications to an input image $x$ based on a textual instruction $c_{T}$, such that the generated output image $y$ semantically aligns with the instruction. For a target image $y$ and an encoder $\mathcal{E}$, the diffusion process adds noise progressively to the encoded latent representation $\mathbf{z}_{0} = \mathcal{E}(y)$, resulting in a noisy latent $\mathbf{z}_t$ with noise level increasing over time steps $t$. A UNet is then trained to predict the injected noise $\epsilon$ in $\mathbf{z}_t$, conditioned on both the image condition $c_x=\mathcal{E}(x)$ and the textual instruction $c_T$. The objective function of latent diffusion is defined as:
\begin{equation}
\ell_{\text {diffusion }}=\mathbb{E}_{\mathcal{E}(y), \mathcal{E}(x), c_{T}, \epsilon \sim N(0,1), t}\left[\| \epsilon-\epsilon_{\theta}\left(t, \text { concat }\left[\mathbf{z}_t, \mathcal{E}(x)\right], c_{T}\right) \|_{2}^{2}\right],
\label{eqdiffusion}
\end{equation}
where $\epsilon$ denotes unscaled noise, $\mathbf{z}_{t}$ is the latent at the sampling step $t$, $\epsilon_\theta$ is the UNet trained to predict the noise added to the noisy latent $\mathbf{z}_{t}$, and $c_T$ is the text condition embedding.

\textbf{IP-Adapter.}
\label{sec:ipadapter}
 IP-Adapter~\cite{ye2023ip} is a lightweight adapter designed to equip pre-trained text-to-image diffusion models with image prompting capability. It injects image features into the diffusion model through decoupled cross-attention modules, enabling simultaneous support for both text and image prompts while preserving a compact architecture.
In the original Stable Diffusion (SD)~\cite{stablediffusion} pipeline, text features extracted by the CLIP~\cite{radford2021learningclipi} text encoder are fed into the cross-attention layers of the UNet to guide image synthesis. To extend this mechanism, the IP-Adapter introduces a decoupled cross-attention scheme, where an additional attention layer is added parallel to the existing text-guided one, specifically for processing image features.
Given query features $\textbf{Z}$, text features $c_T$, and image features $c_I$, the attention operations proceed as follows. The text-guided attention computes $\textbf{Z}' = \text{Softmax}(\textbf{QK}^\top / \sqrt{d})\textbf{V}$, where $\textbf{Q} = \textbf{ZW}_q$, $\textbf{K} = c_T \textbf{W}_k$, and $\textbf{V} = c_T \textbf{W}_v$. Similarly, the image-guided attention computes $\textbf{Z}'' = \text{Softmax}(\textbf{QK}'^\top / \sqrt{d})\textbf{V}'$, where $\textbf{K}' = c_I \textbf{W}_k'$ and $\textbf{V}' = c_I \textbf{W}_v'$. The final fused output is given by $\textbf{Z}_{\text{new}} = \textbf{Z}' + \textbf{Z}''$.
By separating the attention pathways, the model avoids interference from simple feature concatenation and enables more effective fusion of multimodal signals.
In our framework, IP-Adapter serves as the decoder for MLLM-derived visual representations, injecting high-level visual guidance into the diffusion model to steer the image generation process.
\subsection{Overview of MIND-Edit Framework}
An overview of our framework is presented in Fig.~\ref{fig:overview}.
MIND-Edit takes text instruction, the original image, and an editing mask (optional) as inputs and outputs an edited image. MIND-Edit consists of two major collaborative strategies:
\textbf{(1)} instruction optimization strategy: We utilize the rich world knowledge and visual understanding capabilities of a multi-modal large language model (MLLM) to comprehend and reason about the editing intent, thereby optimizing user-provided instruction. This provides more refined guidance for the subsequent image generation process.
\textbf{(2)} MLLM insight-driven image editing strategy: In this part, by training a vision head $\mathbf{F}_{v}$, the intermediate hidden states of the MLLM are taken as input to obtain a visual representation that contains information about the edited image. The visual representation is directly fed into the IP-Adapter, adjusting the cross-attention layers of the diffusion model to guide it in generating an image that conforms to the editing semantics. This directly applies the understanding and reasoning capabilities of the MLLM to the diffusion model. Besides, motivated by recent findings~\cite{tong2024metamorph}, we hypothesize that jointly optimizing the textual and visual outputs of the MLLM can mutually enhance its understanding capabilities across both modalities. Therefore, MIND-Edit adopts a joint training approach using a single multimodal large language model to simultaneously generate optimized text instruction and visual representation. By working together through dual modalities, they promote each other, further stimulating the image understanding ability of the multimodal large language model and its reasoning ability for image editing tasks. Moreover, the visual representations are directly compatible with the IP-Adapter and can be used without training a separate adapter or modifying the diffusion model, enabling a plug-and-play integration into existing generation models.

\begin{figure}[!t]
    \centering
    \includegraphics[width=1.0\textwidth]{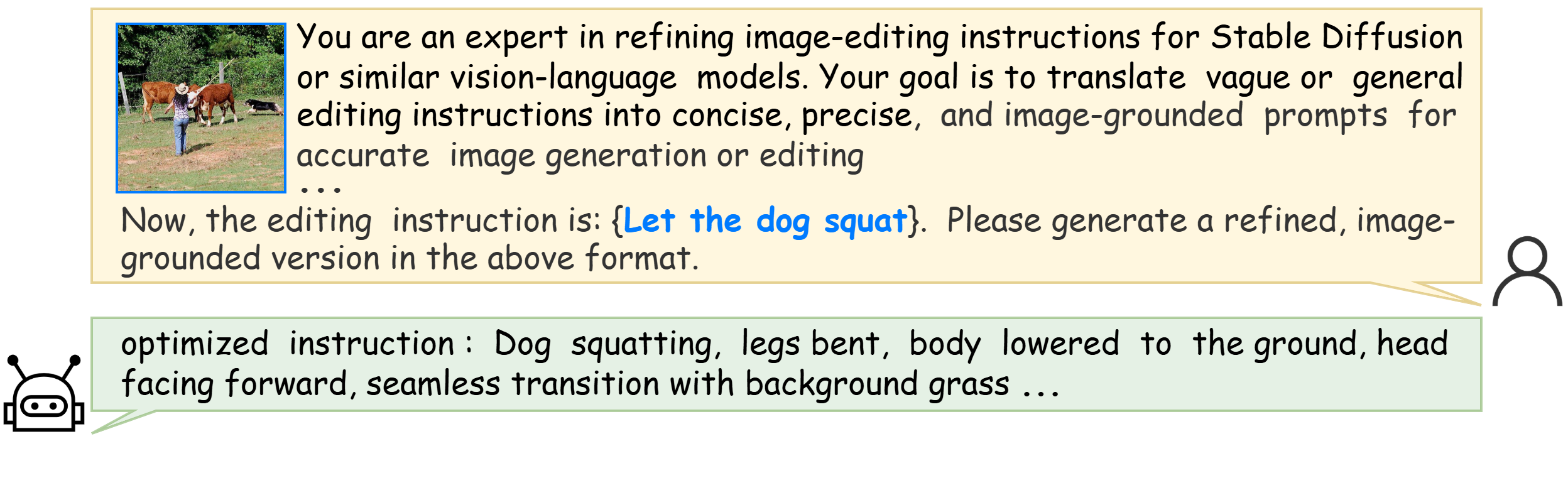}
\caption{\textbf{Illustration of the text instruction optimization strategy.} A prompt informs the MLLM about the upcoming instruction optimization task. Given an image and an instruction from the user, the MLLM refines the instruction by resolving ambiguities based on visual and textual context.}

    \label{fig:prompt}
\end{figure}

\subsection{Text Instruction Optimization Strategy}
User instructions are often vague or underspecified, making it difficult to precisely convey editing intent. To address this issue, MIND-Edit introduces an instruction optimization strategy that refines ambiguous input into clear, executable directives using a multimodal large language model (MLLM). An example of the optimization process is illustrated in Fig.~\ref{fig:prompt}. Through this strategy, users only need to provide a simple text instruction $T_\text{raw}$ along with the image to be edited $I_\text{src}$, and they can obtain an optimized instruction $T_\text{opt}$ that more accurately aligns with their editing intent and is more precise. 
\subsection{MLLM Insight-Driven Image Editing Strategy}
Existing approaches fail to fully leverage the inherent visual understanding capability of multimodal large language models (MLLMs), as they typically use MLLMs to process both textual and visual inputs but ultimately generate intermediate textual outputs to guide the editing process. This reliance on text as the sole modality for downstream control limits the potential of multimodal reasoning. To address this limitation, we propose a novel strategy in which the MLLM directly predicts a visual representation of the desired edited image based on the original image and textual instruction, thereby enabling a more direct connection between multimodal comprehension and visual generation. As shown in Fig.~\ref{fig:overview}, the generated visual representation is injected into the diffusion model via the IP-Adapter, effectively integrating MLLM-derived visual guidance into the image editing pipeline.

\textbf{Visual Representation Generation.}
For the MLLM used in MIND-Edit, the inputs are the textual instruction $T_\text{raw}$ and the image to be edited $I_\text{src}$. The MLLM processes these inputs and produces a sequence of hidden states $\mathbf{h} = \{\mathbf{h}_1, \mathbf{h}_2, ..., \mathbf{h}_N\}$ across $N$ layers. Denoting $\mathbf{F}_{v}$ as the vision head, visual representation $v_{\text{insight}}$ is calculated as:
\begin{equation}
   v_{\text{insight}}= \mathbf{F}_{v}(\mathbf{h}_m),
    \label{vinsight}
\end{equation}
where $\mathbf{h}_m$ denotes the hidden states from the $m$-th layer of the MLLM. Through this design, the MLLM infers the user’s editing intent and transforms it into a visual representation $v_{\text{insight}}$, which encodes the visual information of the desired edited image and provides precise guidance for downstream image editing tasks.


\textbf{Visual Representation Decoding.}
To decode the visual representation $v_{\text{insight}}$ to guide the generation process, MIND-Edit employs the pre-trained IP-Adapter, which equips the diffusion model with image prompting capability by injecting $v_{\text{insight}}$ into its internal feature space, thereby enabling the model to visualize the intended editing outcome.
To be noted, the dimension of $v_{\text{insight}}$ obtained by $\mathbf{F}_{v}$ is directly compatible with the $clip\_embedding$ shape expected by the IP-Adapter. As described in Sec.~\ref{sec:ipadapter}, $v_{\text{insight}}$ is used as $c_I$ and passed into the IP-Adapter to guide the diffusion model. The IP-Adapter injects the MLLM-derived visual representation $v_{\text{insight}}$ into the diffusion model, efficiently bridging the gap between semantic visual intent and the model’s latent feature space, while significantly reducing computational overhead.

\subsection{Joint Training Approach}
Inspired by MetaMorph~\cite{tong2024metamorph}, we adopt a joint training approach in MIND-Edit to unlock and enhance the MLLM’s capabilities for both textual and visual understanding, aiming to improve the quality of instruction optimization and visual representation generation. To this end, MIND-Edit fine-tunes an existing MLLM to simultaneously produce optimized textual instruction $T_\text{opt}$ and visual representation $v_{\text{insight}}$. For the MLLM to be fine-tuned, the inputs are the original image $I_\text{src}$ and the textual instruction $T_\text{raw}$.
Generally, the MLLM uses a language text head $\mathbf{F}_{t}$, which takes the final-layer hidden states $\mathbf{h}_N$ of the MLLM backbone as input and maps them to the vocabulary space to generate a probability distribution over text tokens.  The sequence of tokens with the highest probabilities is then selected as the text output. In this work, for instruction optimization, the original $\mathbf{F}_{t}$ is retained and takes the $\mathbf{h}_N$ as input to produce the optimized textual instruction $T_\text{opt}$, as expressed in Eq.~(\ref{lmhead}):
\begin{equation}
T_\text{opt} = \mathbf{F}_{t}(\mathbf{h}_N).
\label{lmhead}
\end{equation}
The instruction optimization component is trained using the cross-entropy loss, as defined in Eq.~(\ref{eqtextloss}),
\begin{equation}
\ell_\text{text} = -\sum_{t=1}^{n} \log P(w_t \mid w_{<t}, I_\text{src}, T_\text{raw}),
\label{eqtextloss}
\end{equation}
where $n$ is the total number of tokens in the target sequence (i.e., the optimized instruction $T_\text{gt}$ in the training data), $w_t$ is the $t$-th token, and $w_{<t}$ denotes the first $t-1$ ground truth tokens in $T_\text{gt}$. $I_\text{src}$ denotes the input image, and $T_\text{raw}$ is the user-provided textual instruction. The term $P(w_t \mid \cdot)$ represents the probability of the $t$-th token predicted by the model, output by $\mathbf{F}_{t}$ in MIND-Edit. 

To optimize the final visual representation $v_{\text{insight}}$ in MIND-Edit, embeddings extracted from the target edited images $(I_{\text{gt}})$ using the CLIP image encoder~\cite{radford2021learningclipi} are used as ground truth during training. The loss function computes the cosine similarity between the output of the MLLM and the ground truth, expressed as Eq.~(\ref{eqcosloss}):
\begin{equation}
    \ell_{\text{embed}} = 1 - \cos\left(v_{\text{insight}}, \mathcal{E}_{\text{CLIP}}\left(I_{\text{gt}}\right)\right),
    \label{eqcosloss}
\end{equation}
where $\mathcal{E}_{\text{CLIP}}(\cdot)$ denotes the CLIP image encoder, $I_{\text{gt}}$ is the target edited image in the training data, and $\cos(\cdot)$ computes the cosine similarity.
Notably, we use intermediate hidden states $\mathbf{h}_m$ as the input to the vision head $\mathbf{F}_{v}$ to generate $v_{\text{insight}}$, while the final-layer hidden states $\mathbf{h}_N$ are used by the text head $\mathbf{F}_{t}$ to produce the optimized instruction. This separation avoids gradient interference during training, which could otherwise hinder the convergence of the loss function.

Finally, the total loss for joint training is a weighted sum of the text instruction optimization loss and the visual representation loss, formulated as Eq.(\ref{eqtotalloss}).
\begin{equation}
\ell_{\text{total}} = \ell_{\text{text}} + \lambda \ell_{\text{embed}},
\label{eqtotalloss}
\end{equation}
where $\lambda$ is a coefficient that balances the contribution of the two components.
\section{Experiments}\label{sec:experiment}
\subsection{Experimental Settings}
\textbf{Datasets.} A high-quality dataset named HumanEdit~\cite{bai2024humanedit} is used for training and testing. The training set consists of $5244$ randomly selected images, while the test set contains the remaining $500$ images. The dataset covers six types of editing instructions, simulating a wide range of real-world scenarios.
To further evaluate our method in more challenging scenarios, we also conduct experiments on the ComplexMultistepImageEditing\footnote{\url{https://huggingface.co/datasets/NilanE/ComplexMultistepImageEditing}}~\cite{complex-multistep-image-editing-dataset} dataset, which contains $120$ carefully constructed editing examples. The dataset is well-suited for testing the method's capability to handle complex user intent. It is constructed through an iterative refinement process, where outputs are repeatedly evaluated and improved until the editing goal is met, ensuring high quality and consistency.

\textbf{Implementation Details.} We adopt LLaVA-OneVision-7B~\cite{llavaonevision} as our base model. LLaVA-OneVision is trained on single-image, multi-image, and video data in a unified manner, which enables cross-scenario multimodal understanding and robust performance on complex visual tasks. In the joint training approach, the visual representation is obtained by feeding the output of the $21$st hidden layer of LLaVA-OneVision into $\mathbf{F}_{v}$. LLaVA-OneVision has a total of $28$ layers. The training data consists of $5244$ samples from the HumanEdit dataset, and the model is trained for $5$ epochs, with a batch size of $8$ and a learning rate of $1\!\times\!10^{-5}$. During the joint training process, the parameter $\lambda$ was set to $2$.
The diffusion model used is stable‑diffusion‑v1‑5~\cite{stablediffusion}. The text instruction optimization training data is constructed using GPT-4o~\cite{openai2024gpt4ocard}. The prompt used in the construction is shown in Sec.~\ref{sec:prompt-appendix} of the Appendix. Training is completed in approximately $7$ hours using a single NVIDIA A100 GPU.

\begin{table*}[!t]
\caption{\textbf{Quantitative evaluation} of methods on the HumanEdit and ComplexMultistepImageEditing datasets. Best values are highlighted in \best{bold red}, and second-best values are highlighted in \secondbest{underlined blue}. ↑ indicates higher is better; ↓ indicates lower is better. Our method demonstrates competitive performance in both datasets, particularly in handling complex editing tasks.}
\vspace{6pt}
\label{tab:compare_sota}
\centering
\resizebox{1.0\textwidth}{!}{
\begin{tabular}{Sl||ScScScSc||ScScScSc} 
\shline
\rowcolor[HTML]{EFEFEF} 
\multicolumn{1}{c||}{\cellcolor[HTML]{EFEFEF}Method} & 
  \multicolumn{4}{c||}{\cellcolor[HTML]{EFEFEF}HumanEdit~\cite{bai2024humanedit}} & 
  \multicolumn{4}{c}{\cellcolor[HTML]{EFEFEF}ComplexMultistepImageEditing~\cite{complex-multistep-image-editing-dataset}} \\ 
  \hhline{>{\arrayrulecolor[HTML]{EFEFEF}}->{\arrayrulecolor{black}}|--------|}
\rowcolor[HTML]{EFEFEF} 
\cellcolor[HTML]{EFEFEF} & 
  \cellcolor[HTML]{EFEFEF}CLIP-I $\uparrow$ & 
  \cellcolor[HTML]{EFEFEF}LPIPS $\downarrow$ & 
  \cellcolor[HTML]{EFEFEF}PSNR $\uparrow$ & 
  \cellcolor[HTML]{EFEFEF}SSIM $\uparrow$ & 
  \cellcolor[HTML]{EFEFEF}CLIP-I $\uparrow$ & 
  \cellcolor[HTML]{EFEFEF}LPIPS $\downarrow$ & 
  \cellcolor[HTML]{EFEFEF}PSNR $\uparrow$ & 
  \cellcolor[HTML]{EFEFEF}SSIM $\uparrow$ \\

\hline
SmartEdit~\cite{huang2024smartedit}  & 0.8841 & 0.2915 & 17.1728 & 0.6828 & 0.5096 & 0.7630 & 7.7464 & 0.1333 \\
BrushNet~\cite{ju2024brushnet}       & 0.8986 & 0.1830 & 19.2172 & 0.7877 & \secondbest{0.6924} & 0.7464 & 8.1696 & 0.1281 \\
MagicQuill~\cite{liu2024magicquill} & \best{0.9381} & \best{0.1162} & \secondbest{22.2380} & \best{0.8981} & \best{0.6996} & \secondbest{0.7037} & \secondbest{8.8158} & \secondbest{0.1817} \\
MIND-Edit \textbf{(Ours)}                                & \secondbest{0.9310} & \secondbest{0.1245} & \best{22.2714} & \secondbest{0.8517} & 0.6340 & \best{0.6963} & \best{8.9245} & \best{0.2069} \\
\shline
\end{tabular}}
\end{table*}

\textbf{Evaluation Metrics.} For the edited images, we conduct quantitative experiments using four metrics: CLIP-I~\cite{radford2021learningclipi}, LPIPS~\cite{zhang2018unreasonablelpips}, PSNR, and SSIM~\cite{wang2004imagessim}, which assess performance across three dimensions: semantic alignment, perceptual quality, and structural fidelity. 

\subsection{Comparisons with State-of-the-Art Methods}

We conduct two sets of comparative experiments to comprehensively evaluate the performance of different image editing methods. The first set focuses on simple editing tasks and is evaluated on the HumanEdit dataset~\cite{bai2024humanedit}, using $500$ randomly selected test images not included in training. The second set targets complex editing scenarios and is conducted on the ComplexMultistepImageEditing dataset~\cite{complex-multistep-image-editing-dataset}. The compared methods include SmartEdit~\cite{huang2024smartedit}, BrushNet~\cite{ju2024brushnet}, and MagicQuill~\cite{liu2024magicquill},  all of which incorporate MLLMs into their editing pipelines. SmartEdit and BrushNet are built on stable‑diffusion‑v1‑5, while MagicQuill adopts a dual-branch architecture with one branch based on stable‑diffusion‑v1‑5 and the other on stable-diffusion-inpainting.
\begin{figure}[!t]
    \centering
    \includegraphics[width=1\textwidth]{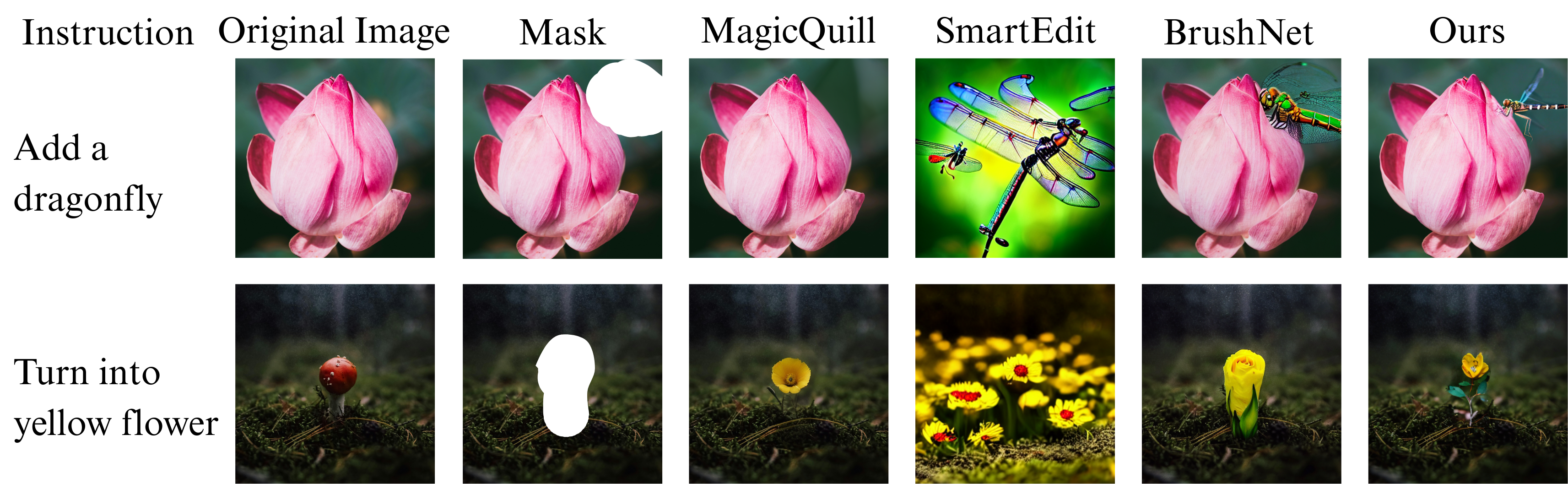}
    \caption{\textbf{Qualitative comparisons on the HumanEdit dataset}~\cite{bai2024humanedit}. A mask is provided for each sample. MIND-Edit achieves superior instruction alignment and visual quality, surpassing or matching other methods even though MagicQuill's generation branch alone contains twice the parameters of our method.}
    \label{fig:compare}
\end{figure}
\begin{figure}[t]
    \centering
    \includegraphics[width=1\textwidth]{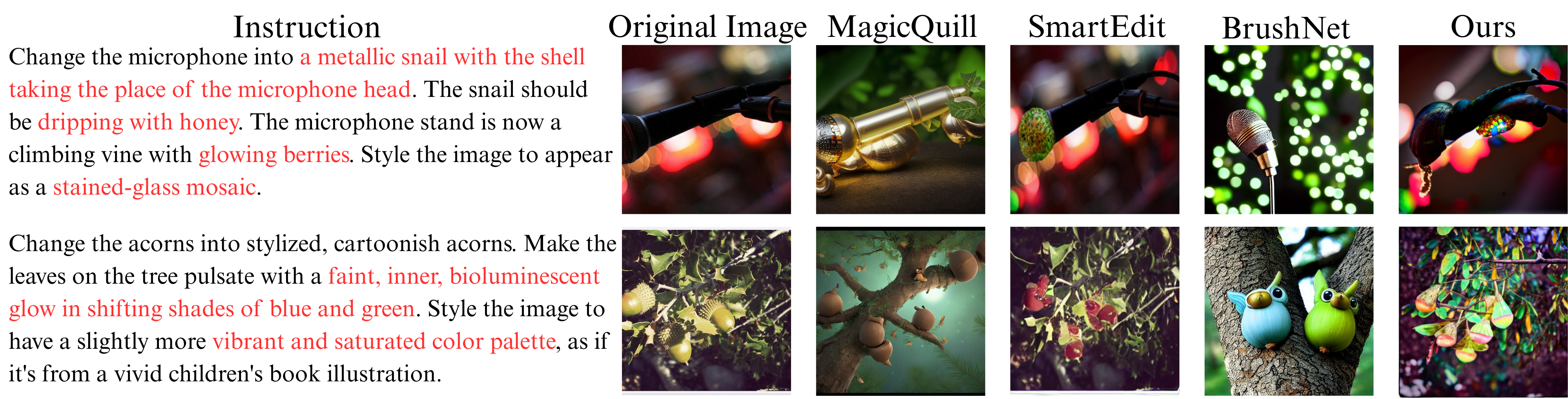}
    \caption{\textbf{Qualitative comparisons on the ComplexMultistepImageEditing dataset}~\cite{complex-multistep-image-editing-dataset}, where no mask is provided. MIND-Edit achieves precise semantic alignment in complex editing scenarios.}
    \label{fig:complex-compare}
\end{figure}
The quantitative results are summarized in Table~\ref{tab:compare_sota}. In simple editing scenarios, MIND-Edit significantly outperforms SmartEdit and BrushNet across quantitative metrics and performs comparably to MagicQuill, despite the latter having twice as many parameters. In complex editing scenarios, MIND-Edit surpasses the MagicQuill method in most metrics and is far ahead of SmartEdit and BrushNet, which shows that MIND-Edit has significant advantages in handling challenging tasks.
Fig.~\ref{fig:compare} illustrates that MIND-Edit accurately executes the intended edits, successfully adds a dragonfly, and transforms mushrooms into yellow flowers. It also delivers superior visual details, particularly in the dragonfly and the green leaves of the flower, compared to other methods. For more qualitative comparisons, please refer to Sec.~\ref{sec:compare-appendix} of the Appendix.
Fig.~\ref{fig:complex-compare} further shows that MIND-Edit achieves precise semantic alignment in a mask-free manner under complex editing scenarios.  Specifically, the shape and orientation of the snail and the layout of the dripping honey in the first group of edits all meet the intent of the instruction. The color and tone of the acorns in the second group all met the instruction description, and the density and layout of the acorns were largely consistent with the original image. Across both cases, the background and overall tone more accurately reflect the instruction semantics than those generated by other methods.

\subsection{Ablation Study}

\textbf{Effect of proposed Strategies.}
To verify that both instruction optimization and visual representation guidance contribute to better image editing performance, the following ablation experiments are conducted:
\textbf{\#$1$}: Use only the optimized instruction obtained by the joint training approach without $v_{\text{insight}}$.
\textbf{\#$2$}: Use only $v_{\text{insight}}$ obtained by joint training approach without instruction optimization.
The results are displayed in the first two lines of Table~\ref{tab:ablation_unified}.

\textbf{Effect of Joint Training Approach.}
To demonstrate that the joint training approach can unlock the intrinsic textual and visual understanding capabilities of the MLLM, we independently fine-tune two separate models based on the pre-trained LLaVA-OneVision-7B: one for instruction optimization and the other for visual representation generation. We then compare these independently trained models with the jointly trained model proposed in this work.
Recall that for generating visual representations, the $21$st-layer hidden states of the backbone network are used as input to $\mathbf{F}_{v}$ to obtain $v_{\text{insight}}$, while for instruction optimization, the final-layer hidden states are fed into $\mathbf{F}_{t}$ to produce the optimized textual instruction.
To validate the effectiveness of the joint training approach in improving image editing performance, we conduct four experiments:
\textbf{\#$3$}: Both $T_\text{opt}$ and $v_{\text{insight}}$ are obtained through independent training.
\textbf{\#$4$}: $T_\text{opt}$ is obtained through joint training, while $v_{\text{insight}}$ is obtained through independent training.
\textbf{\#$5$}: $T_\text{opt}$ is obtained through independent training, while $v_{\text{insight}}$ is obtained through joint training.
The results are reported in rows $3$ to $5$ of Table~\ref{tab:ablation_unified}.
\textbf{\#$6$ (Ours)} represents the full MIND‑Edit method, where both the optimized instruction and $v_{\text{insight}}$ are obtained through the joint training approach.

\begin{table*}[t]
  \centering
  \caption{\textbf{Results of ablation study.} "$\checkmark$" in the Strategy columns indicates that the corresponding strategy is enabled, whereas a "—" means it is disabled. 
In the Training Approach columns, "I"  denotes using the independent training approach, and "J" denotes using the joint training approach.}
  \label{tab:ablation_unified}
  \small
  \vspace{6pt}
\resizebox{1.\linewidth}{!}{
\begin{tabular}{Sl||ScSc||ScSc||ScScScSc}
\shline
\rowcolor[HTML]{EFEFEF}
\multicolumn{1}{c||}{\cellcolor[HTML]{EFEFEF}Exp.\#} &
  \multicolumn{2}{c||}{\cellcolor[HTML]{EFEFEF}Strategy} &
  \multicolumn{2}{c||}{\cellcolor[HTML]{EFEFEF}Training Approach} &
  \multicolumn{4}{c}{\cellcolor[HTML]{EFEFEF}Metrics} \\
\rowcolor[HTML]{EFEFEF}
\cellcolor[HTML]{EFEFEF} &
  \cellcolor[HTML]{EFEFEF}$T_\text{opt}$ &
  \cellcolor[HTML]{EFEFEF}$v_{\text{insight}}$ &
  \cellcolor[HTML]{EFEFEF}$T_\text{opt}$ &
  \cellcolor[HTML]{EFEFEF}$v_{\text{insight}}$ &
  \cellcolor[HTML]{EFEFEF}CLIP‑I $\uparrow$ &
  \cellcolor[HTML]{EFEFEF}LPIPS $\downarrow$ &
  \cellcolor[HTML]{EFEFEF}PSNR $\uparrow$ &
  \cellcolor[HTML]{EFEFEF}SSIM $\uparrow$ \\
\shline
\#$1$ & $\checkmark$ & — & J & — &
  \secondbest{0.9307} & 0.1264 & \secondbest{22.4354} & 0.8433 \\
\#$2$ & — & $\checkmark$ & — & J &
  0.9287 & 0.1281 & 22.0378 & 0.8491 \\
\hline
  \#$3$ & $\checkmark$ & $\checkmark$ & I & I &
  0.9009 & 0.1549 & 19.9760 & 0.8283 \\
\#$4$ & $\checkmark$ & $\checkmark$ & J & I &
  0.9145 & 0.1306 & 22.2412 & 0.8507 \\
\#$5$ & $\checkmark$ & $\checkmark$ & I & J &
  0.9275 & \secondbest{0.1260} & \best{22.3914} & \best{0.8524} \\
\#$6$ (Ours) & $\checkmark$ & $\checkmark$ & J & J &
  \best{0.9310} & \best{0.1245} & 22.2714 & \secondbest{0.8517} \\
\shline
\end{tabular}}
\end{table*}
\begin{figure}[!t]
    \centering
    \includegraphics[width=1.0\textwidth]{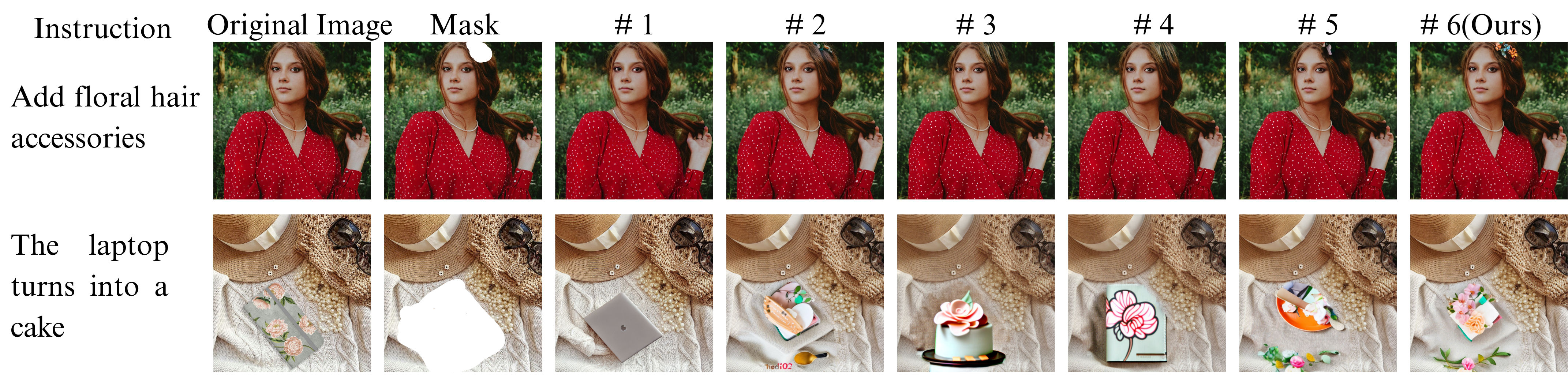}
   \caption{\textbf{Qualitative results of the ablation study.} With the proposed instruction optimization, visual representation generation strategies, and joint training approach, MIND-Edit achieves improved instruction-aligned details, textures, and overall visual consistency compared to other variants.}
    \label{fig:A}
\end{figure}

The results of experiments \textbf{\#$1$}, \textbf{\#$2$}, and \textbf{\#$6$ (Ours)} in Table~\ref{tab:ablation_unified} indicate that both the instruction optimization and the use of visual representation contribute to improved image editing performance. The results of experiments \textbf{\#$3$}, \textbf{\#$4$}, \textbf{\#$5$}, and \textbf{\#$6$ (Ours)} indicate that both the instruction optimization strategy and the MLLM insight-driven image editing strategy benefit from the joint training approach, with consistent improvements across all quantitative metrics.
The qualitative results of the ablation study are presented in Fig.~\ref{fig:A}. In the results of the \textbf{\#$1$}, the target object specified in the instruction does not appear, indicating that guidance from $v_{\text{insight}}$
can improve the accuracy of editing tasks. In \textbf{\#$2$}, the generated images contain artifacts, suggesting that optimized instructions help improve image quality. Compared to the \#$6$, \#$3$, \#$4$, and \#$5$ experiments exhibit poorer detail, texture, and overall accuracy, demonstrating the effectiveness of the joint training approach. For more qualitative results of the ablation study, please refer to Sec.~\ref{sec:ablation-appendix} of the Appendix.
\section{Conclusion}
We propose MIND-Edit, a framework that enables high-precision editing in complex visual scenarios. By combining instruction optimization with visual representations predicted by the MLLM to guide the editing process, our method effectively captures user intent, accurately identifies the regions to be edited, and produces results aligned with user expectations.  Extensive experiments across both simple and complex scenarios demonstrate that MIND-Edit consistently performs well under diverse and challenging conditions. MIND-Edit contributes innovative strategies to the image editing field and offers promising directions for future research. In future work, we plan to validate our framework on broader datasets and explore its applicability in more practical and diverse scenarios.

\textbf{Limitations.} Although MIND-Edit provides an effective and theoretically plug-and-play image editing framework, the current experimental results are obtained using the Stable Diffusion model and may therefore be constrained by its generative quality. Additionally, the visual representation is injected via the IP-Adapter, whose limited capacity may also restrict performance. Employing more advanced diffusion models or stronger adapter architectures could potentially lead to further improvements in editing quality.
\bibliographystyle{plain}
\bibliography{reference}

\clearpage
\appendix
\setcounter{section}{0}
\setcounter{table}{0}
\renewcommand\thetable{\Alph{table}}
\makeatletter
  \renewcommand\theHtable{\Alph{table}}   
\makeatother
\setcounter{figure}{0}
\renewcommand\thefigure{\Alph{figure}}
\makeatletter
  \renewcommand\theHfigure{\Alph{figure}}
\makeatother

\section{More Results of Comparative Experiments}
\label{sec:compare-appendix}
\begin{figure}[ht]
    \centering
\includegraphics[width=1.0\textwidth]{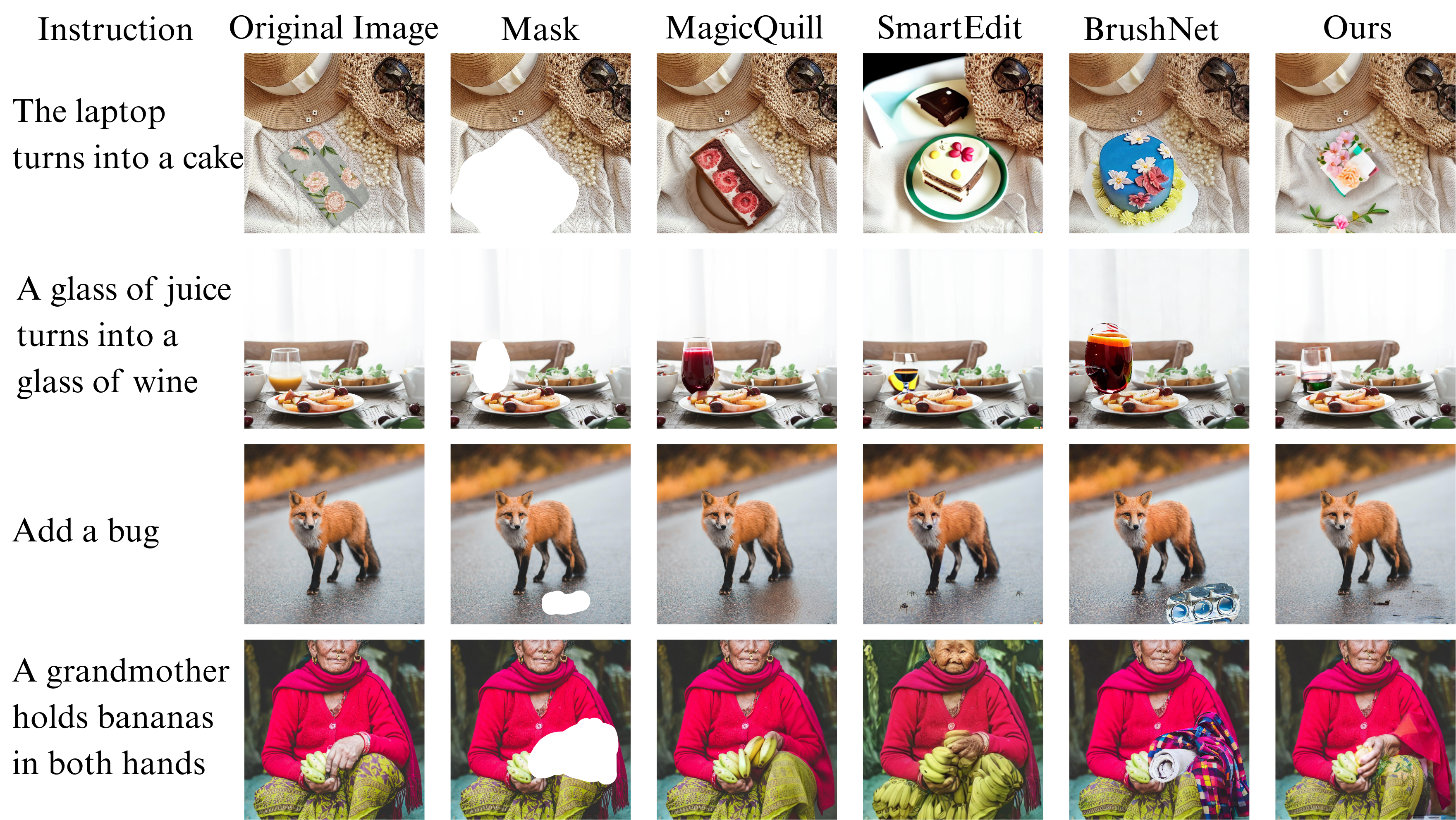}
    \caption{Qualitative comparison of MIND-Edit with SmartEdit, BrushNet, and MagicQuill on the HumanEdit dataset.}
    \label{fig:compare-appendix}
\end{figure}

Fig.~\ref{fig:compare-appendix} presents additional qualitative results on the HumanEdit dataset, comparing MIND-Edit with SmartEdit, BrushNet, and MagicQuill. All methods are evaluated using the same experimental settings as described in Sec.~\ref{sec:experiment}. Notably, although MagicQuill adopts a dual-branch architecture, which employs a generation module with twice the parameter count of our model, MIND-Edit still achieves comparable or superior visual performance. The examples further highlight MIND-Edit’s superior performance in executing editing instructions.
\begin{figure}[!t]
    \centering
\includegraphics[width=1.0\textwidth]{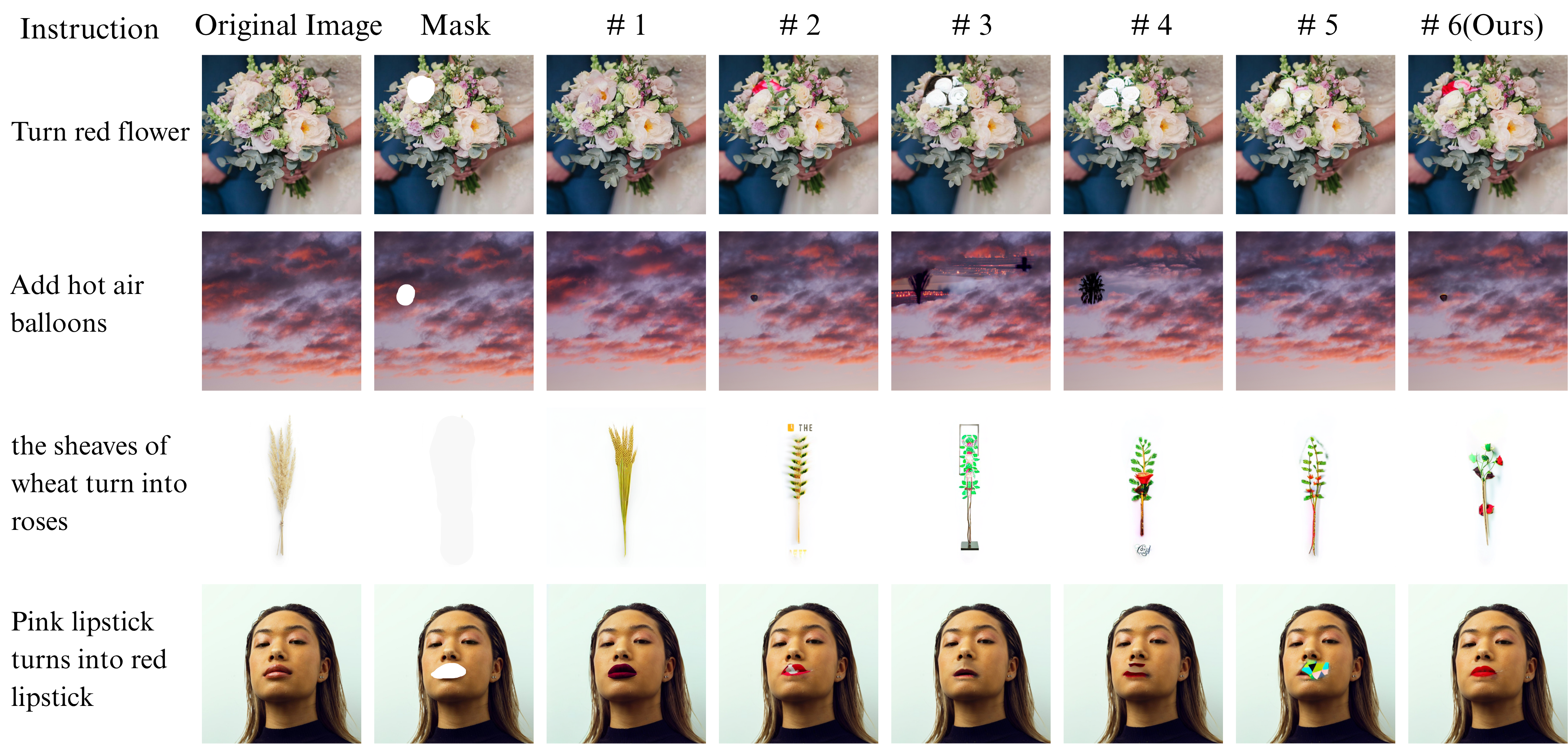}
    \caption{Qualitative results of the ablation study on the HumanEdit dataset.}
    \label{fig:ablation-appendix}
\end{figure}
\begin{figure}[!t]
    \centering
\includegraphics[width=1.0\textwidth]{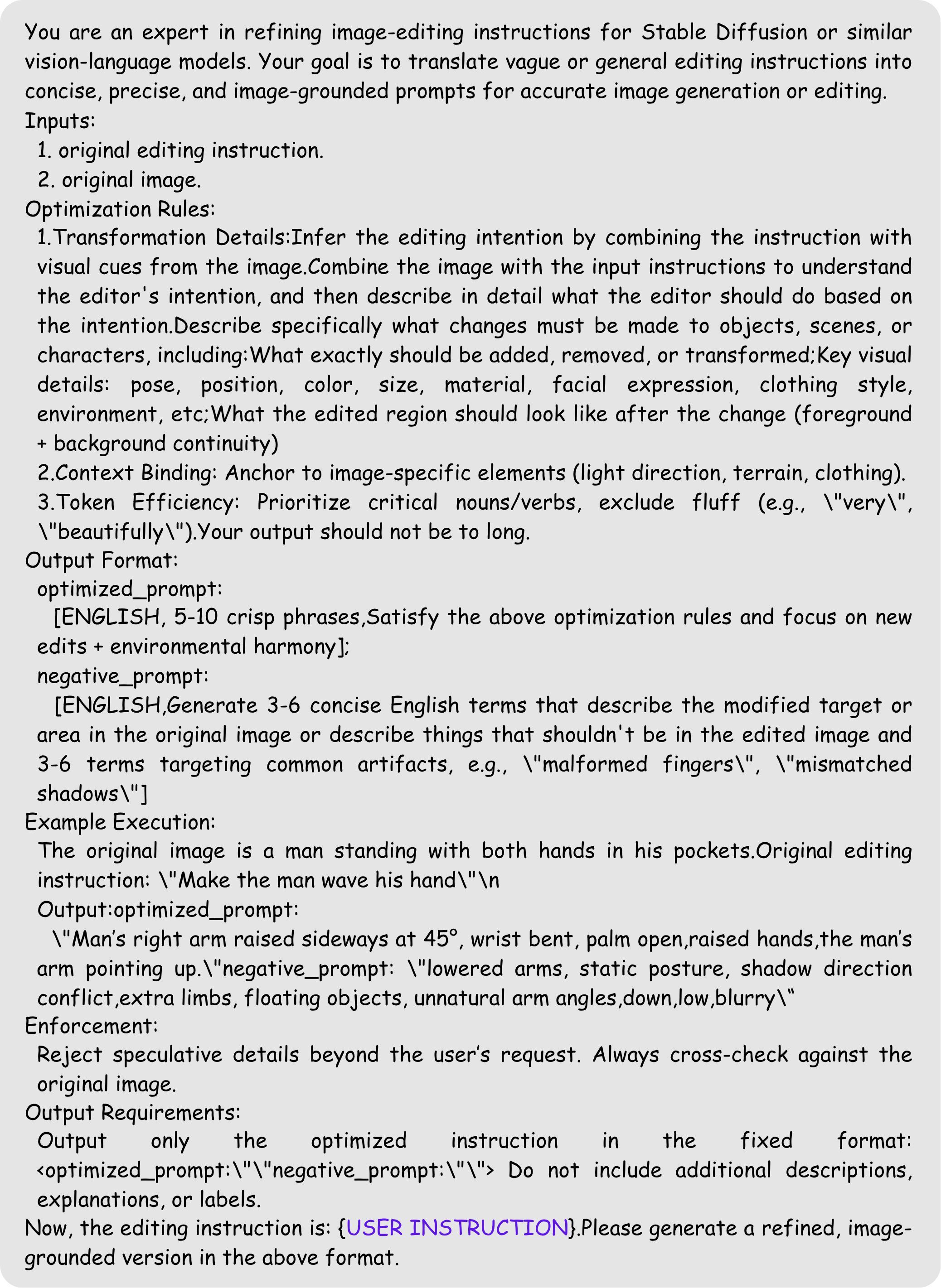}
    \caption{Prompt used with GPT-4o to construct optimized instruction training data.}
    \label{fig:prompt-appendix}
\end{figure}

\section{More Results of Ablation Study}
\label{sec:ablation-appendix}
Fig.~\ref{fig:ablation-appendix} provides additional qualitative results from the ablation study conducted on the HumanEdit dataset. The experimental settings are consistent with those described in Sec.~\ref{sec:experiment}. These examples further illustrate the individual contributions of instruction optimization and visual representation, as well as the effectiveness of the joint training approach adopted in MIND-Edit.
\section{Designed Prompts}
\label{sec:prompt-appendix}
Fig.~\ref{fig:prompt-appendix} shows the prompt used with GPT-4o to construct high-quality training data for the instruction optimization component of MIND-Edit. This prompt guides GPT-4o in transforming vague, underspecified user instructions into precise, executable directives grounded in the image content. The purple-highlighted section labeled USER INSTRUCTION indicates the original instruction to be optimized.

The prompt includes three core elements: (1) Transformation Rules for inferring user intent by combining image and instruction, specifying object-level changes such as pose, position, and appearance; (2) Context Binding to align prompts with image-specific cues like lighting and layout; (3) Token Efficiency to emphasize critical content while removing unnecessary phrases.
The output follows a fixed format with an optimized\_prompt ($5$–$10$ concise phrases) and a negative\_prompt (terms to suppress artifacts or conflicts), ensuring the data is both compact and informative.
This constructed supervision data enables the MLLM to better understand user intent and generate high-fidelity editing outputs.
\end{document}